# Research on Autonomous Maneuvering Decision of UCAV Based on Approximate Dynamic Programming

Zhencai Hu, Peng Gao, Fei Wang

Harbin Institute of Technology, Shenzhen

## ABSTRACT

Unmanned aircraft systems can perform some more dangerous and difficult missions than manned aircraft systems. In some highly complicated and changeable tasks, such as air combat, the maneuvering decision mechanism is required to sense the combat situation accurately and make the optimal strategy in real time. This paper presents a formulation of a 3-D one-on-one air combat maneuvering problem and an approximate dynamic programming approach for computing an optimal policy on autonomous maneuvering decision making. The aircraft learns combat strategies in a Reinforcement Leaning method, while sensing the environment, taking available maneuvering actions and getting feedback reward signals. To solve the problem of dimensional explosion in the air combat, the proposed method is implemented through feature selection, trajectory sampling, function approximation and Bellman backup operation in the air combat simulation environment. This approximate dynamic programming approach provides a fast response to a rapidly changing tactical situation, learns in a long-term planning, without any explicitly coded air combat rule base.

## 1. INTRODUCTION

Unmanned combat aerial vehicle (UCAV) is a highly efficient military aircraft that carries kinds of weapons and perform their missions automatically, UCAVs have been successfully employed to replace manned aircrafts in a variety of military aerial missions for their high mobility and large overloaded maneuvers. It is foreseeable that UCAV will become the primary choice in the future air combat field. Autonomous maneuvering decision is a mechanism with respect to how the UCAV choose tactical plans or maneuvers during the process. The performance of this mechanism plays a crucial role in the complex and constantly changing air-to-air combat missions. The goal of autonomous maneuvering in air combat games is to develop a method that can make effective maneuvering decisions online, and compute desirable sequentially maneuvers without directly expert pilot inputs.

Air combat has been explored by several researchers in the past. The autonomous maneuvering decision methods for UCAV can be divided into two categories. One is the non-learning method, such as influence diagram[1,2], differential game[3] and expert system[4]. There is no training and optimizing process in the non-learning strategies. Moreover, these maneuvering strategies are fixed and limited for situational awareness, and it is difficult to completely cover all air combat situations. The other category is the self-learning method, including genetic algorithm[5], artificial immune system[6] and reinforcement learning[7], and so forth. These methods adopt their own experiences, and the models are optimized to cope with complex and changeable environments. The reinforcement learning (RL) methods have been successfully applied to solve many difficult decision making tasks[8,9].

Unfortunately, in the large-scale air combat problems, RL-based dynamic programming (DP) method are intractable because of the problem of exponential dimensional explosion, which indicates that it is impossible to recode information over all the continuous air combat states. Function approximation[10] is capable of producing superior results in a finite time by approximating state value functions or state-action value functions.

In this paper, we create a 3-D air combat adversarial simulation environment and train an automatic online maneuvering decision model based on the approximate dynamic programming (ADP) method. First, a simplified combat environment is modeled to simulate the dynamic process. Then a learning mechanism based on ADP method is built. Finally, the effectiveness of automatic maneuvering decision-making strategy is verified by simulation test.

## 2. APPROXIMATE DYNAMIC PROGRAMMING TO AIR COMBAT

Given a model of UCAV dynamics and an approximate objective function, the dynamic programming provides the efficient means to precisely compute an optimal maneuvering policy for the air combat game. The optimal policy provides the best sequentially actions at any state, eliminating the massive online computation. This section gives a brief review of the dynamic programming, and describes the proposed approximate solution in detail.

### 2.1 Dynamic Programming

There are some terminologies will be used in the DP method. The UCAV air combat simulation is a process of continuous decision making on UCAV, which can be abstracted into a Markov decision process[11] (MDP). State vector $s_i$ represents the combat situation at each state $i = 1, \ldots, n$, and the whole state space is consisted of all the single discrete states as $S = [s_1 \ s_2 \ \cdots \ s_n]$. The UCAV agent is allowed to take actions $a_i$ from action set $A = [a_1 \ a_2 \ \cdots \ a_N]$ at each step. And a scalar feedback reward signal $r$ is released from the environment. Given the current state and action input, we use the state transition function $f(s, a)$ to compute the next state based on the dynamics of UCAV movement.

The optimal policy is accomplished by the optimal future reward value of each state $V^*(s)$. A value function $V(s)$ is defined at each state, representing the long-term future reward from this state, which is used to evaluate the goodness or badness of the state. This optimal long-term reward function can be computed by executing the value iteration, which is repeatedly performing a Bellman backup[12] operation on each state, defined as

$$V^{k+1}(s) = \max_{a \in A} \left\{ r_s^a + \gamma V^k[f(s,a)] \right\} \quad (1)$$

where $\gamma \in [0,1]$ is the discount factor applied at each one-step look forward prediction, and $r_s^a$ is the immediate reward by taking action $a$ at state $s$. The Bellman backup optimization can be implemented on many states simultaneously. After adequate iteration, $V(s)$ will converge to the optimal value $V^*(s)$. Then an optimal policy $\pi^*$ can be determined from $V^*(s)$. By playing greedy strategy[13,14], the optimal action at each time step is defined as

$$a_i = \pi^*(s_i) = \arg\max_{a \in A} \left\{ r_{s_i}^a + \gamma V^*[f(s_i,a)] \right\} \quad (2)$$

### 2.2 Function approximation

A continuous function approximator eliminates the need to represent and compute the future reward for every available state $s_i$ [15]. Some correlated features are chosen to approximate the value function of a high-dimensional state space. The components of state vector $s_i$ can take any continuous value within the boundary. The original discrete look-up table of values $V(s)$ is now replaced by the approximated value $V_{approx}(s)$ which is initialized to be zero at all locations. And the state space $S$ can be determined by some manageable state sampling approach. The Bellman backup operation is applied to each state sample, and the iteration values are stored in target vector $\hat{V}^{k+1}(S)$:

$$\hat{V}^{k+1}(S) = \max_{a \in A} \left\{ r_S^a + \gamma V_{approx}^k[f(S,a)] \right\} \quad (3)$$

The linear estimator uses a set of descriptive features $\phi(s)$ and the target vector to estimate the future reward function $V_{approx}^{k+1}(s)$ by standard least-square estimation.

$$V_{approx}^{k+1}(s) = \phi(s)(\Phi^T \Phi)^{-1} \Phi^T \hat{V}^{k+1}(s) \quad (4)$$

where $\Phi = [\phi_1 \ \phi_2 \ \cdots \ \phi_m]^T$ stores the features set computed from all available states $s_i \in S$. The ADP architecture relieves some of the difficulty associated with dimensional curse in classical DP techniques.

## 3. AIR COMBAT ENVIRONMENT DESIGN

In this section, the system states, control actions, aircraft dynamics, goal and reward function are described for the specific one-on-one air combat simulation.

### 3.1 State, Actions and Dynamics

Assuming the aircraft as a particle, the combat state is defined by the speed, position and the flight attitude angles of a red UCAV (denoted by the $r$ subscript) and a blue UCAV (denoted by the $b$ subscript) at any step. The state vector is

$$s = [v_r, x_r, y_r, z_r, \theta_r, \psi_r, \varphi_r, v_b, x_b, y_b, z_b, \theta_b, \psi_b, \varphi_b] \tag{5}$$

where $x$, $y$ and $z$ indicate the coordinates of the craft in the three-dimensional spatial inertial system, $x$ represents the east axis, $y$ represents the north axis and $z$ represents the height axis. $v$ represents the flying velocity, $\theta$, $\psi$ and $\varphi$ refer to the pitch angle, the yaw angle and the roll angle of the aircraft, respectively.

According to the kinematic principles, the flight dynamics of both sides follow the differential equations shown as formula (6).

$$\begin{aligned}
\dot{v} &= g(N_x - \sin\theta) \\
\dot{\psi} &= gN_z \sin\varphi / (v\cos\theta) \\
\dot{\theta} &= (N_z \cos\varphi - \cos\theta)g/v \\
\dot{x} &= v\cos\theta \sin\psi \\
\dot{y} &= v\cos\theta \cos\psi \\
\dot{z} &= v\sin\theta
\end{aligned} \tag{6}$$

$N_x$ is the tangential overload, $N_z$ is the normal overload and $\varphi$ is the roll angle of the aircraft. Relevantly, the list of $[N_x, N_z, \varphi]$ is regarded as the control commands input to the system. The dynamics explain the state transition mechanism. Given the current state values, successively UCAV maneuvers and a suitable simulation time interval $\Delta t$, the following states can be solved with the state transaction function $f(s,a)$ using the fourth order Runge-Kutta[16] method. The state samples for ADP training process are sampled by the trajectory sampling to ensure that the areas most likely to be seen during combat were sampled sufficiently. The simulation was initialized again at a randomly generated state. This process continued until all needed points were generated.

The establishment of the UCAV maneuver library can refer to the pilot commands. NASA have designed seven typical flight maneuver models that can be used in air combat, which are continued flight, maximum acceleration, maximum deceleration, turn left, turn right, pull up and push down. The corresponding control inputs $[N_x, N_z, \varphi]$ of these maneuvers are shown in Table 1.

Table 1. Control inputs for basic UCAV maneuver Commands

| Maneuvering inputs | Air Combat Maneuvers Library | | | | | | |
|---|---|---|---|---|---|---|---|
| | continued | Acceleration | Deceleration | Turn left | Turn right | Pull up | Push down |
| $N_x$ | 0 | 2 | 0 | 0 | 0 | 0 | 0 |
| $N_z$ | 1 | 1 | 1 | 5 | 5 | 5 | -5 |
| $\varphi$ | 0 | 0 | 0 | $-\pi/3$ | $\pi/3$ | 0 | 0 |

### 3.2 Goal and Reward function

The goal of UCAV is to attain and maintain a position of advantage over the other aircraft and shot the enemy aircraft, while minimizing the risk to its own body. The reward function decides the direction of model optimization and is determined through the real-time combat situation features at every step. Figure 1 shows the spatial relative relationships between the two aircrafts.

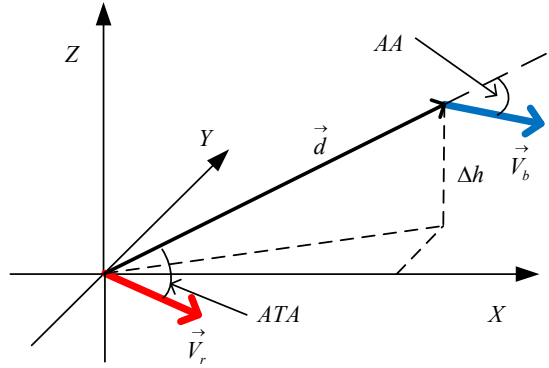

Figure1 Spatial relative relationship between the red and blue aircraft

The angle factors in air combat are mainly Aspect Angle (AA) and Antenna Train Angle (ATA). From the perspective of the red aircraft, the AA represents the angle between the distance vector and the tail of the enemy's speed vector. The ATA is the angle between the red speed vector and the radar guideline. Both AA and ATA can be calculated using the opponents of the state vector as formula (7).

$$\begin{aligned}
\vec{v}_r &= [\cos\psi_r \cos\theta_r \quad \sin\psi_r \cos\theta_r \quad \sin\theta_r] \\
\vec{v}_b &= [\cos\psi_b \cos\theta_b \quad \sin\psi_b \cos\theta_b \quad \sin\theta_b] \\
\vec{d} &= [x_r - x_b \quad y_r - y_b \quad z_r - z_b] \\
AA &= \arccos(\vec{v}_b \cdot \vec{d} / |\vec{d}|) \\
ATA &= \arccos(\vec{v}_r \cdot \vec{d} / |\vec{d}|)
\end{aligned} \qquad (7)$$

The extracted features include velocity advantage reward $\Delta v = v_r - v_b$, height advantage reward $\Delta h = z_r - z_b$, and the distance-angle reward, as Eq. 8. All the three rewards are scaled to the interval $[-1,1]$, and the total reward function at every step is the weighted sum of them, as Eq.9.

$$R_3 = \left[\frac{(1-|AA|/\pi)+(1-|ATA|/\pi)}{2}\right] \bullet e^{-(\frac{|d|-R_d}{k\pi})} \qquad (8)$$

$$R = (R_1 + R_2 + R_3)/3 \qquad (9)$$

The reward function for the blue UCAV can be computed in the same way. The feature vector chosen to approximate the value function is $\phi(s) = [AA, ATA, \Delta z, \Delta v, |\vec{d}|]$.

## 4. EXPERIMENTS

As we have talked all the environment design and ADP application in air combat. In this section we perform simulations and demonstrate the capability of UCAV to make intelligent successively maneuvering decisions.

We set the red and the blue UCAV flying oppositely. The original positional component values are sampled from a Gaussian distribution with standard deviation $\sigma = 10$ m, and means $\mu$ listed in Table 2. The pitch angle and yaw angle are sampled from a narrow range ($\pm 3°$) of uniform distribution around the values in Table 2. The states space in the ADP model includes 100,000 discrete samples obtained by trajectory sampling, and the number of dynamic programming learning iterations is set as 40.

Table 2. Initial state values used for simulation

| $S\_init$ | $v_r$ (m/s) | $x_r$ (m) | $y_r$ (m) | $z_r$ (m) | $\theta_r$ (°) | $\psi_r$ (°) | $v_b$ (m/s) | $x_b$ (m) | $y_b$ (m) | $z_b$ (m) | $\theta_b$ (°) | $\psi_b$ (°) |
|---|---|---|---|---|---|---|---|---|---|---|---|---|
| values | 250. | 0. | 0. | 2900. | 0. | 45. | 204. | 3000. | 3000. | 2800. | 0. | -135. |

The time interval in the whole simulation process is $\Delta t = 0.25$s, that is, every maneuvering decision is made after 0.25s. Once a UCAV enter the dominated area of the opponent enemy or the total number of steps in current episode reaches a certain predefined maximum number, the combat episode ends immediately. The dominated area is defined as $|ATA| < 1.1$ and $|AA| < 0.6$, and the maximum simulation length of each air combat episode is defined as 200 to avoid excessive spatial scales in the 3-D simulation.

We present two combat situations. In the first case, the red side UCAV makes online maneuvering decisions according to the policy learned from ADP method, while the blue side UCAV keeps the continued flight maneuvering and flies in a straight line. As we can see in Figure 2, when the blue side flies straightly, the red UCAV learns an effective tactic strategy that pull up the aircraft and gains the advantage of height firstly, and then turn the direction around and dive to the back side of the enemy aircraft immediately to obtain the relative dominative area. In the second case, we let both the red side and blue side UCAV make maneuvering decisions through the ADP decision model, which is also known as a self-play process. Figure 3 shows some typical combat episodes.

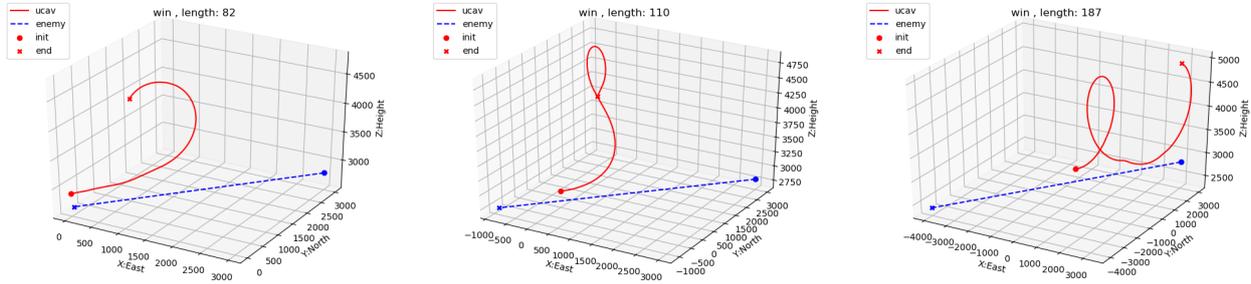

Figure 2. Typical combat result against continued flying opponent

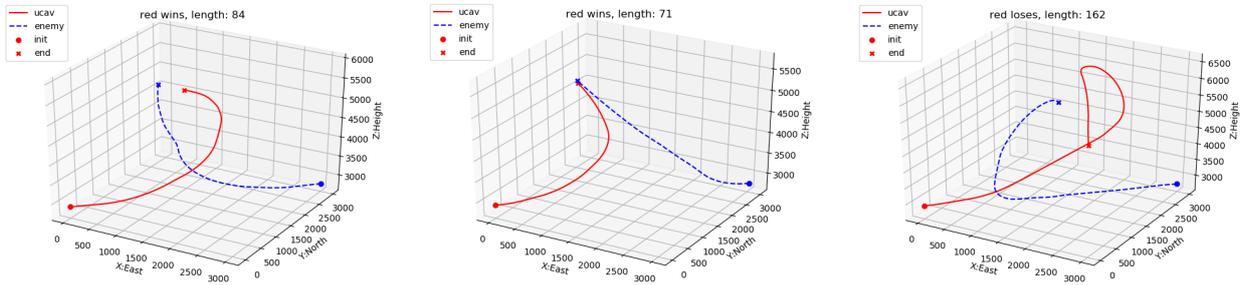

Figure 3. Typical combat result of self-play fight

## 5. CONCLUSION

In this paper, we successfully apply the approximate dynamic programming method to the air combat which avoids the problem of dimensional explosion problem by leading in the function approximation and dynamic programming method.

Specifically, we create a simplified 3-D one-on-one adversarial air combat simulation environment, design the UCAV available maneuvers, and specify the reward function of the UCAV which directs the optimization direction. Moreover, a training and testing model is constructed. Without the involvement of experienced air combat pilot, the UCAV can intelligently implement an effective tactic strategy after a certain amount of training. Experimental evaluations demonstrate that our agent UCAV can learn the offensive and defensive tactics in the highly flexible air combat game. Due to the limitation of the computing power, we constrain the air combat scene in a relatively simplified occasion and decide the action choice as discrete values. The continuous action value maneuvering model would be more complicated and likely to the real pilot maneuvering. In the future, more factors can be considered, such as expanding the spatial scales of air combat, adding different weapons, and even many-to-many UCAV battle scenarios. The air combat problem is still a highly complex issue. With the development of intensive learning technology and artificial general intelligence, both flighting and decision-making capabilities of UCAV will reach a new level of intelligence.